\documentclass[10pt,twocolumn,letterpaper]{article}

\usepackage{cvpr}
\usepackage{times}
\usepackage{epsfig}
\usepackage{graphicx}
\usepackage{amsmath}
\usepackage{amssymb}
\usepackage{algorithmicx,algorithm}
\usepackage[noend]{algpseudocode}
\usepackage{array, bm, placeins, lipsum}
\usepackage{booktabs, color, enumerate, microtype, multirow, subcaption}
\usepackage[table]{xcolor}
\newcolumntype{x}[1]{>{\centering\arraybackslash\hspace{0pt}}p{#1}}

\usepackage[pagebackref=true,breaklinks=true,letterpaper=true,colorlinks,bookmarks=false]{hyperref}

\cvprfinalcopy 

\def\cvprPaperID{1640} 

\ifcvprfinal\fi
\begin{document}

\title{Recurrent Residual Module for Fast Inference in Videos}

\author{Bowen Pan$^{1\dag}$, Wuwei Lin$^{1\dag}$, Xiaolin Fang$^{2\P}$, Chaoqin Huang$^{1\dag}$, Bolei Zhou$^{3\S}$, Cewu Lu$^{1\ddag}$\thanks{Cewu Lu is the corresponding author.} \\
\small{$^1$Shanghai Jiao Tong University, $^2$Zhejiang University, $^3$Massachusetts Institute of Technology}\\
{\tt\small $^\dag$\{googletornado,linwuwei13, huangchaoqin\}@sjtu.edu.cn, $^\P$fxlfang@gmail.com}\\ 
{\tt\small $^\S$bzhou@csail.mit.edu; $^\ddag$lu-cw@cs.sjtu.edu.cn}
}

\maketitle

\begin{abstract}
	Deep convolutional neural networks (CNNs) have made impressive progress in many video recognition tasks such as video pose estimation and video object detection. However, CNN inference on video is computationally expensive due to processing dense frames individually. In this work, we propose a framework called \textit{Recurrent Residual Module} (RRM) to accelerate the CNN inference for video recognition tasks. This framework has a novel design of using the similarity of the intermediate feature maps of two consecutive frames, to largely reduce the redundant computation. One unique property of the proposed method compared to previous work is that feature maps of each frame are precisely computed. The experiments show that, while maintaining the similar recognition performance, our RRM yields averagely 2$\times$ acceleration on the commonly used CNNs such as AlexNet, ResNet, deep compression model (thus 8$-$12$\times$ faster than the original dense models using the efficient inference engine), and impressively 9$\times$ acceleration on some binary networks such as XNOR-Nets (thus 500$\times$ faster than the original model). We further verify the effectiveness of the RRM on speeding up CNNs for video pose estimation and video object detection. 
\end{abstract}

\vspace{-0.4cm}
\section{Introduction}\label{intro}
Video understanding is one of the long-standing topics in computer vision. Recently, deep convolutional neural networks (CNNs) advanced different tasks of video understanding, such as video classification~\cite{KarpathyCVPR14,zha2015exploiting,Ng_2015_CVPR,zhou2017temporal}, video pose estimation~\cite{fang2016rmpe, cao2016realtime}, and video object detection~\cite{girshick2014rich, girshick2015fast, ren2015faster, liu2016ssd, redmon2016you, redmon2016yolo9000}. However, using CNNs to process the dense frames of videos is computationally expensive while it becomes unaffordable as the video goes longer. Meanwhile, millions of videos are shared on the Internet, where processing and extracting useful information remains a challenge. With the video datasets becoming larger and larger \cite{sigurdsson2017actions, abu2016youtube, KarpathyCVPR14, kay2017kinetics, caba2015activitynet,monfort2018moments}, training and evaluating neural networks for video recognition are more challenging. For example, for Youtube-8M dataset~\cite{abu2016youtube} with over 8 million video clips, it will take 50 years for a CPU to extract the deep features using a standard CNN model. 

One of the bottlenecks for video understanding using CNNs is the frame-by-frame CNN inference. A one-minute video contains thousands of frames thus the model inference becomes much slower in comparison with processing a single image. However, different from a set of independent images, consecutive frames in a video clip are usually similar. Thus, the high-level semantic feature maps in the deep convolutional neural networks of the consecutive frames will also be similar. Intuitively, we can leverage the frame similarity to reduce some redundant computation in the frame-by-frame video CNN inference. An attractive recursive schema is as follows:
\vspace{-0.2cm}
\begin{equation}\label{naive-schema}
\vspace{-0.2cm}
	\mathcal{R}(I_{t}) =  \mathcal{R}(I_{t-1} ) + \mathcal{G}(I_{t} - I_{t-1}),
\end{equation}
where $\mathcal{R}$ is the deep CNN feature, $\mathcal{G}$ is a fast and shallow network that only processes the frame difference between frame $I_t$ and $I_{t-1}$ in a video clip. Ideally, $\mathcal{G}$ should be both efficient and accurate to extract the residual feature. However, it remains challenging to implement such a schema due to the nonlinearity of CNNs.

Some previous works have tried to address this nonlinearity. Zhu \emph{et al.}~\cite{zhu2016deep} proposed deep feature flow framework which utilizes the flow field to propagate the deep feature maps. However, these estimated feature maps will cause a drop on performance compared to the original feature maps. Kang \emph{et al.}~\cite{kang2017optimizing} developed a NoScope system to perform the fast binary query of the absence of a specific category. It is fast but not generic enough for other video recognition tasks.

We propose the framework of Recurrent Residual Module (RRM) to thoroughly address the nonlinear issue of CNNs in Eq.~\ref{naive-schema}. The nonlinearity of CNNs results from the pooling layers and activation functions, while the computationally expensive layers such as convolution layer and fully-connected layer are linear. Thus for two consecutive frame inferences, if we are able to share the overlapped calculation of these linear layers, a large amount of the computation can be eliminated. To this end, we snapshot the input and output feature maps of convolution layers and fully-connected layers for the inference on the next frame. Consequently, we only need to forward pass the frame difference region with the feature maps of the previous frame in each layer, which leads to the sparsity matrix multiplication that can be largely accelerated by the EIE techniques~\cite{Han:2016:EEI:3007787.3001163}. In general, our RRM can dramatically reduce the computation cost from the convolution layers and fully-connected layers, while still maintains the nonlinearity of the whole network.

The main contribution of this work is the framework of Recurrent Residual Module, which is able to speed up almost any CNN-based models for video recognition without extra training cost. To the best of our knowledge, this is the first acceleration method that can compute the feature maps precisely when deep CNNs process videos. We evaluate the proposed method and verify its effectiveness on accelerating CNNs for video recognition tasks such as video pose estimation and the video object detection.

\section{Related Work}
We have a brief survey on the related work of improving the neural network efficiency as below.

\textbf{Network weight pruning.} It is known that removing the redundant model parameters reduces the computational complexity of networks \cite{NIPS1989_250, hanson1989comparing, hassibi1993second, strom1997phoneme, collins2014memory}. At the very beginning, Hanson \& Pratt \cite{hanson1989comparing} applied the weight decay method to prune the network, then Optimal Brain Damage (OBD) \cite{NIPS1989_250} and Optimal Brain Surgeon (OBS) \cite{hassibi1993second} pruned the parameters using the Hessian of the loss function. 
Recently, Han \emph{et al.} \cite{han2015learning, han2015deep} showed that they could even reduce the model parameters by an order of magnitude in deep \emph{CNN} models while maintaining the performance. They devised an efficient inference engine~\cite{Han:2016:EEI:3007787.3001163} to speed up the models. Instead of pruning model weights, our RRM framework focuses on factorizing the input at each layer, then further speeds up the model based on the pruning methods.

\textbf{Network quantization.} Quantizing network weight is to replace the high-precision float numbers of the weights with several limited integers, such as +1/-1~\cite{soudry2014expectation, courbariaux2015binaryconnect, courbariaux2016binarized, rastegari2016xnor, li2017performance} or +1/0/-1~\cite{arora2014provable}. Rastegari \emph{et al.}~\cite{rastegari2016xnor} proposed XNOR-Networks that use both binary weights and binary inputs to achieve 58$\times$ faster convolution operations on a CNN trained on ImageNet. Yet, applying these quantization methods requires retraining the model and also results in a loss of accuracy. 

\textbf{Low rank acceleration.} Decomposing weight tensor based on low-rank methods are used to accelerate deep convolutional networks. Both \cite{denton2014exploiting, jaderberg2014speeding} reduced the redundancy of the weight tensors through the low-rank approximation. Yang \emph{et al.} \cite{yang2015deep} showed that they can use a single Fastfood layer to replace the FC layer. Liu \emph{et al.} \cite{liu2015sparse} reduced the computation complexity using a sparse decomposition. All of these methods speed up the test-time evaluation of convolutional networks with some sacrifice in precision. 

\textbf{Filter optimization.} Reducing the filter redundancy in convolution layers is an effective method to simplify the CNN models~\cite{luo2017thinet,he2017channel,howard2017mobilenets}. Luo \etal~\cite{luo2017thinet} pruned filters and set the output feature maps as the optimization objective to minimize the loss of information. Howard \etal~\cite{howard2017mobilenets} developed MobileNet which applied depth-wise separable convolution to decompose a standard convolution operation and showed an effectiveness. He \etal~\cite{he2017channel} proposed an iterative algorithm to jointly learn additional filters for filter selection and scalar masks for each output channel. They achieved 13$\times$ speedup on AlexNet.

\textbf{Sparsity.} It is most related to our method. Obviously, sparsity can significantly accelerate the convolutional networks both in training and testing \cite{liu2015sparse, changpinyo2017power, guo2016dynamic, wen2016learning}. There are many previous works showing that they can save the energy \cite{chen2017eyeriss, reagen2016minerva} and accelerate the convolution \cite{albericio2016cnvlutin, Shi:2017ug, dong2017more} by skipping the zeros or elements close to zero in the sparse input. Albericio \emph{et al.} \cite{albericio2016cnvlutin} proposed an efficient convolution accelerator utilizing the sparsity of inputs, while Shi \& Chu \cite{Shi:2017ug} sped up the convolution on CPUs by eliminating the zero values in the output of ReLUs. Graham \& Maaten \cite{DBLP:journals/corr/GrahamM17, BenGraham:2015ue} introduced a sparse convolution that eliminated the computation of values in some inactive output positions by recognizing the input cells in the ground state. Recently, Han \emph{et al.}~\cite{Han:2016:EEI:3007787.3001163} devised an efficient inference engine (EIE) that can exploit the dynamic sparsity of the input feature maps to accelerate the inference. Our RRM integrates EIE as a step to further optimize the model weight.

Our Recurrent Residual Module works in a recurrent manner. The most similar architecture to ours is the Predictive-Corrective Networks~\cite{dave2017predictive}, which derives a series of recurrent neural networks to make prediction about feature and then correct them with some bottom-up observations. The key difference, also the most innovative point of our model, is that we utilize the recurrent framework to accelerate CNN models using sparsity and Efficient Inference Engine, which is much more efficient than the Predictive-Corrective Networks~\cite{dave2017predictive}. Besides our method is a generic framework that could be plugged in a variety of CNN models without retraining to speed up the forward pass. 

\begin{figure*}
\begin{center}
\includegraphics[width=\linewidth]{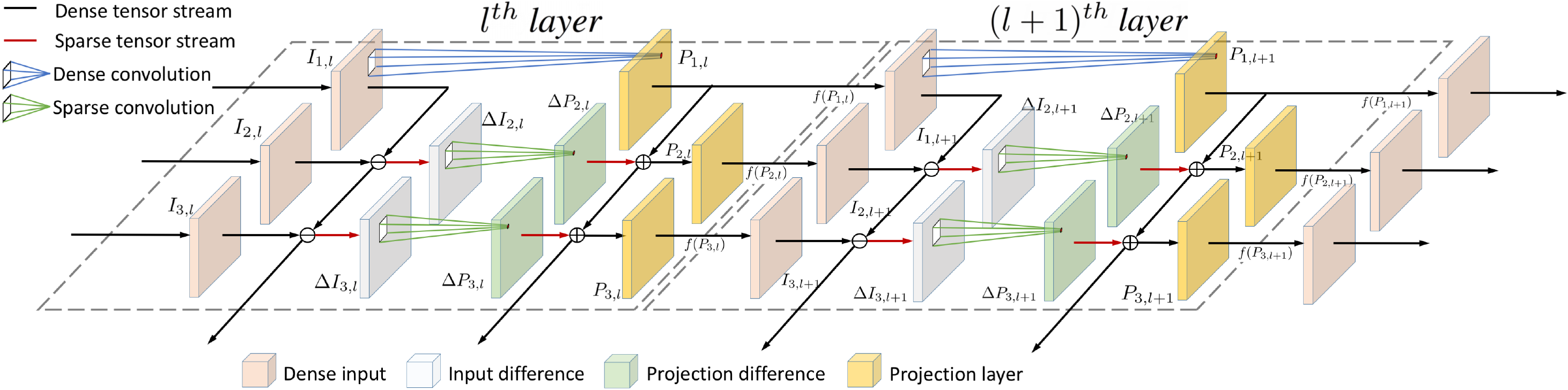}
\end{center}
\vspace{-0.5cm}
   \caption{Illustration of the Recurrent Residual Module on two layers. \emph{Dense convolution} operation represents the standard convolution operation. \emph{Sparse convolution} operation represents the SPMV-equipped convolution operation which will deliver speed up to the sparse input. \emph{Sparse convolution} has no bias term and shares the same weight filters with \emph{Dense convolution}. Mapping $f$ represents some nonlinear operator combinations in CNNs such as ReLUs and max poolings.}
\vspace{-0.2cm}
\label{rra}
\end{figure*}

\vspace{-0.2cm}
\section{Recurrent Residual Module Framework}\label{method}

The key idea of the Recurrent Residual Module is to utilize the similarity between the consecutive frames in a video clip to accelerate the model inference. To be more specific, we first improve the sparsity of the input to each \textbf{linear layer} (layers with linearity, including convolution layer and FC layer), then use the sparse matrix-vector multiplication accelerators (SPMV) to further speed up the forward pass. 

We will first introduce some preliminary concepts and discuss the linearity of convolution layers and FC layers. Then the recurrent residual module will be introduced in detail, followed by the analysis of computation complexity, sparsity enhancement, and accumulated error. Last but not least, we integrate the efficient inference engine \cite{Han:2016:EEI:3007787.3001163} (EIE) to further improve the framework's efficiency.

\subsection{Preliminary}

We denote a standard neural network using the notion set $\langle\mathcal{I},\mathcal{F},*,\mathcal{W}, f\rangle$, where $\mathcal{I}$ represents the set of input tensor (it could be the input image or the output from the previous layer), $\mathcal{F}$ is the set of weight filters in convolution layers, $*$ denotes the convolution operations, $\mathcal{W}$ represents the set of weight tensors in FC layers, and $f$ represents some nonlinear operators.
In convolution phase, $f$ can be a ReLU~\cite{icml2010_NairH10} or a pooling operator. And in the fully-connected phase, it can be a short-cut function.

We use $I_{tl}\in\mathcal{I}$ to denote the input tensor to the $l^{th}$ linear layer when we process the $t^{th}$ frame in the video, $W_{l}\in\mathcal{W}$ to represent the weight tensor of the $l^{th}$ layer if it is FC layer, $F_{l}\in\mathcal{F}$ to represent the weight filter of the $l^{th}$ layer if it is convolution layer. When processing the $t^{th}$ frame, the $l^{th}$ layer performs the following operation:
\vspace{-0.2cm}
\begin{equation}\label{operation1}
\vspace{-0.2cm}
	I_{t(l+1)} =
	\left\{
	\begin{aligned}
	& f(F_{l}*I_{tl} + b_{l}) & \footnotesize{\emph{if $l^{th}$ layer is convolution}} \\
	& f(W_{l}I_{tl} + b_{l}) & \footnotesize{\emph{if $l^{th}$ layer is FC}}\\
	\end{aligned}
	\right. ,
\end{equation}
where $b_l$ is the bias term of the $l^{th}$ layer.
And we define the \textbf{projection layer} $P_{tl}$ as:
\vspace{-0.2cm}
\begin{equation}\label{operation2}
\vspace{-0.2cm}
	P_{tl} =
	\left\{
	\begin{aligned}
	& F_{l}*I_{tl} + b_{l} & \footnotesize{\emph{if $l^{th}$ layer is convolution}} \\
	& W_{l}I_{tl} + b_{l} & \footnotesize{\emph{if $l^{th}$ layer is FC}}\\
	\end{aligned}	
	\right. .
\end{equation}
Due to the linearity of convolution operation and multiplication operation, given the difference of $P_{tl}$ and $P_{(t-1)l}$, we have:
\vspace{-0.2cm}
\begin{equation}\label{operation3}
\vspace{-0.2cm}
	P_{tl} - P_{(t-1)l} =
	\left\{
	\begin{aligned}
	& F_{l}*\Delta I_{tl} & \footnotesize{\emph{if $l^{th}$ layer is convolution}} \\
	& W_{l}\Delta I_{tl} & \footnotesize{\emph{if $l^{th}$ layer is FC}}\\
	\end{aligned}	
	\right. ,
\end{equation}
where $\Delta I_{tl}=I_{tl}-I_{(t-1)l}$. Thus Eq.~\ref{operation1} can be written as:
\vspace{-0.2cm}
\begin{equation}\label{operation4}
\vspace{-0.1cm}
	I_{t(l+1)} =
	\left\{
	\begin{aligned}
	& f(P_{(t-1)l}+F_l*\Delta I_{tl} ) &  \footnotesize{\emph{if $l^{th}$ layer is convolution}} \\
	& f(P_{(t-1)l}+W_{l}\Delta I_{tl}) & \footnotesize{\emph{if $l^{th}$ layer is FC}} \\
	\end{aligned}	
	\right. .
\end{equation}
Eq.~\ref{operation4} is the key point in our RRM framework. $P_{(t-1)l}$ has been obtained and preserved during the inference phase of the last frame. Evidently, the computation mainly falls on $F_l*\Delta I_{tl}$ or $W_{l}\Delta I_{tl}$. Due to the similarity between the consecutive frames, $\Delta I_{tl}$ is usually highly sparse (This will be verified in our experiment). As a result, to obtain the final result, we just need to work on a rather sparse tensor $\Delta I_{tl}$ instead of the original one $I_{tl}$, which is dense and computationally expensive. With the help of sparse matrix-vector multiplication accelerators (SPMV), the calculations of zero elements can be skipped, thus inference speed is improved.

\subsection{Recurrent Residual Module for Fast Inference}
The illustration of the recurrent residual module (RRM) is shown in Fig.~\ref{rra}. In order to preserve the information of the last frame and obtain the efficient $\mathcal{G}$ which is introduced in Sec.~\ref{intro}, the information of input tensor to each linear layer $\mathcal{I}_{(t-1)}$ and the corresponding projection layer set $\mathcal{P}_{(t-1)}$ of each linear layer is saved. The preserved information can be applied during the inference phase for the following frame.

As shown in the Fig.~\ref{rra}, in the inference stream of frame~$2$, when the input tensor $I_{2,l}$ is fed to the convolution layer (the $l^{th}$ layer), we first subtract $I_{1,l}$ from $I_{2,l}$ to obtain $\Delta I_{2,l}$, where $I_{1,l}$ is the input tensor to the $l^{th}$ layer of frame~$1$ and was snapshotted when processing frame~$1$. As illustrated in the previous discussion, $\Delta I_{2,l}$ is a sparse tensor. Apply the sparse matrix-vector multiplication accelerator to the $l^{th}$ layer, we can skip the zero elements and get the convolution result within a short time. Next, the output of the convolution layer is snapshotted. Add the output to projection layer $P_{1,l}$, we can obtain the intact tensor that is exactly the same as the output of a normal convolution layer which is fed $I_{2,l}$. After that, we perform the nonlinear mapping $f$ to $P_{1,l}$. In this manner, the final result is obtained. To some extent, it is similar to the distributive law of multiplication.

The specific procedure of the inference with Recurrent Residual Module is listed in Algorithm~\ref{algorithm1}.
\begin{algorithm}
\caption{Inference with Recurrent Residual Module}
\label{algorithm1}
\hspace*{0.02in} {\textbf{Input}:}
A video clip $\mathcal{X}:=\{ x_t|t=1,2,...,T \}$,where $x_t$ is the frame at time $t$, a pre-trained neural network $\mathcal{M}$. \\
\hspace*{0.01in} {\textbf{Output}:}
Frame-level feature set $\mathcal{F}:=\{ f_t|t=1,2,...,T\}$, where $f_t$ is the deep feature of the frame $x_t$.
\begin{algorithmic}[1]
\State{$I_{0,:} \Leftarrow 0$}
\For{$t=1$ to $T$}
		\State{$I_{t,0}\Leftarrow x_t$}
		\For{$l,L$ in \emph{enumerate($\mathcal{M}$)}}
		\State{$\hat{I}_{t,l}\Leftarrow I_{t,l}-I_{(t-1),l}$}
		\If{$L$ is \emph{convolution layer}}
			\State{$P_{tl} = P_{(t-1)l} + F_l*\hat{I}_{t,l}$}
		\Else
			\State{$P_{tl} = P_{(t-1)l} + W_l\hat{I}_{t,l}$}
		\EndIf
		\State{$I_{t,l+1}\Leftarrow f(P_{tl})$}
		\EndFor
		\State{$f_t\Leftarrow I_{t,T}$}
\EndFor
\end{algorithmic}
\end{algorithm}

One drawback of the RRM is that we can only forward pass frames with the help of the feature snapshots of the previous frames, which limits doing inference in parallel for the whole video. To address this we can split the video into several chunks then process each chunk with RRM-equipped CNN in parallel. 

\subsection{Analyzing computational complexity}
\renewcommand \arraystretch{1.2}
\begin{table}[!htb]
  \centering
  \small
  \vspace{-0.3cm}
  \scalebox{1}{
	\begin{tabular}{ccx{3.3cm}x{3.3cm}}
    \toprule
    Layer Type & Complexity \\
    \cmidrule(lr){1-1} \cmidrule(lr){2-2}
    Convolution layer & $O(W_{c_i}H_{c_i}C_{c_i}^{in}C_{c_i}^{out}w_Fh_F)$ \\
    \cellcolor{gray!15}Convolution layer + SPMV & \cellcolor{gray!15}$O(\rho_{c_i}W_{c_i}H_{c_i}C_{c_i}^{in}C_{c_i}^{out}w_Fh_F)$   \\
    \cmidrule(lr){1-1} \cmidrule(lr){2-2}
    FC layer & $O(C_{f_j}^{in}C_{f_j}^{out})$ \\
    \cellcolor{gray!15}FC layer + SPMV & \cellcolor{gray!15}$O(\rho_{f_j}C_{f_j}^{in}C_{f_j}^{out})$  \\
    \bottomrule
  \end{tabular}
  }
  \vspace{-0.2cm}
  \caption{Ablation analysis of computational complexity. Layers equipped with SPMV will skip the calculations of zero elements.}
  \vspace{-0.2cm}
  \label{complexity}
\end{table}
The computational complexity of the neural network with the recurrent residual module in test-phase is analyzed. In a sequence of convolution layers $M_{c_1},M_{c_2},...,M_{c_n}$, suppose that for layer $M_{c_i}$, the density (the proportion of non-zero elements) of the input tensor $I_{c_i}\in R^{C_{c_i}^{in}\times W_{c_i}\times H_{c_i}}$ is $\rho_{c_i}$, the weight matrices is $F_{c_i}\in R^{C_{c_i}^{in}\times C_{c_i}^{out} \times W_F \times H_F}$. Similarly, for an FC layer $M_{f_j}$, we have the density $\rho_{f_j}$, the input vector $I_{f_j}\in R^{C_{f_j}^{in}}$ and the weight tensor $W_{f_j}\in R^{C_{f_j}^{in}\times C_{f_j}^{out}}$.

In our Recurrent Residual Module, compared to the multiplication operation, both execution time and computational cost of add operation are trivial. Hence, to analyze the computation complexity, the following discussion will only focus on the multiplication complexity in the original linear layer and in our RRM framework. Table~\ref{complexity} shows the multiplication complexity of a single layer. For the entire neural network, the computational complexity after utilizing the sparsity can be calculated as follows (assume that the stride is $1$):
\vspace{-0.2cm}
\begin{equation}\label{eq:speed-up-ratio}
\vspace{-0.3cm}
	O(\sum_{i}\rho_{c_i}W_{c_i}H_{c_i}C_{c_i}^{in}C_{c_i}^{out}w_Fh_F+\sum_{j}\rho_{f_j}C_{f_j}^{in}C_{f_j}^{out}).
\end{equation}
Eq.~\ref{eq:speed-up-ratio} illustrates that the sparsity (the proportion of zero elements) of the input tensor to each layer is the key to reduce the computation cost. In terms of the sparsity, some networks equipped with ReLU activation functions already have many zero elements in their feature maps. In our recurrent residual architecture, the sparsity can be further improved as discussed below.

\subsection{Improving sparsity}\label{sparse_enhance}

Our framework can obtain the inference output identical to the original model without any approximation. And we could further improve the sparsity of the intermediate feature map to approximate the inference output as a trade-off to further accelerate inference. However, it would possibly lead to the issue of error accumulation over time. To address this issue, we estimate the accumulated error given by accumulated truncated values. First, the accumulated truncated values are obtained by
\vspace{-0.3cm}
\begin{equation}
\vspace{-0.4cm}
	e_t = \sum_{t}\sum_{j}\ell_{2}(u_{t,j}),
\end{equation}
where $u_{t,j}$ is the truncated map to the $j^{th}$ linear layer in the inference stream of $t^{th}$ frame. We denote accumulated accuracy error by 
\vspace{-0.3cm}
\begin{equation}
\vspace{-0.2cm}
    e_c = \mathcal{H}(e_t, \mu).
 \end{equation}
$\mathcal{H}$ is a fourth order Polynomial function regression with the parameter $\mu$, which is fitted from large amount of data pairs of accumulated truncate value and accumulated error. If it is larger than a certain threshold, a new precise inference will be carried out to clear accumulated error and a new round of fast inference will start.

\subsection{Efficient inference engine}
To implement the RRM framework efficiently, we utilize dynamic sparse matrix-vector multiplication(DSPMV) technique. While there are a number of existing off-the-shelf DSPMV techniques~\cite{Han:2016:EEI:3007787.3001163,Shi:2017ug}, the most efficient one among them is the efficient inference engine (EIE) proposed by Han \emph{et al.}~\cite{Han:2016:EEI:3007787.3001163}.
	
EIE is the first accelerator which exploits the dynamic sparsity in the matrix-vector multiplications. When performing multiplication between matrix $W$ and sparse vector $a$, the vector $a$ is scanned and a Leading Non-zero Detection Node (LNZD Node) is applied to recursively look for the next non-zero element $a_j$. Once found, EIE broadcasts $a_j$ along with its index $j$ to the processing elements (PEs) which hold the weight tensor in the CSC format. Then weights column $W_j$ with the corresponding index $j$ in all PEs will be multiplied by $a_j$ and the results will be summed into the corresponding row accumulator. These accumulators finally output the resulting vector $b$.

Since the multiplication between matrix and matrix can be decomposed into several matrix-vector multiplication processes, by decomposing the input tensor to several dynamically sparse vectors, we embed the EIE to our RRM framework conveniently.

\vspace{-0.2cm}
\section{Experiments}\label{exp}
\vspace{-0.2cm}
In this section, we first verify that our recurrent residual module can consistently improve the sparsity of the input tensor to each layer in Sec.~\ref{sparse} across different network architectures. We measure the overall sparsity of the whole network to estimate the improvement. The overall sparsity is calculated as the ratio of zero-value elements in the inputs of all linear layers, which is:
\vspace{-0.2cm}
\begin{equation}
\vspace{-0.1cm}
    S = \frac{\sum_{i}s_{c_i}W_{c_i}H_{c_i}C_{c_i}^{in}C_{c_i}^{out}w_Fh_F+\sum_{j}s_{f_j}C_{f_j}^{in}C_{f_j}^{out}}{\sum_{i}W_{c_i}H_{c_i}C_{c_i}^{in}C_{c_i}^{out}w_Fh_F+\sum_{j}C_{f_j}^{in}C_{f_j}^{out}},
\end{equation}
where $s_{c_i}$ and $s_{f_i}$ are the sparsity of the input tensor to the convolution layer $M_{c_i}$ and the FC layer $M_{f_j}$ respectively. Then, we show the speed and accuracy trade-off in our RRM framework. After that, we combine our RRM framework with some classical model acceleration techniques such as the XNOR-Net~\cite{rastegari2016xnor} and the Deep Compression models~\cite{han2015deep} to further accelerate the model inference. Finally, we demonstrate that we can accelerate several off-the-shelf CNN-based models, here we take the detectors in the field of pose estimation and object detection for examples. In this section, we provide a theoretical speedup ratio by computing the theoretical computational time of the EIE~\cite{Han:2016:EEI:3007787.3001163}, which is calculated by dividing the total workload GOPs by the peak throughput. The actual computation time is around $10\%$ more than the theoretical time due to the load imbalance. Yet, this bias will not affect our speedup ratio. For an uncompressed model, EIE has an impressive processing power of 3 TOP/s. We utilize its feature that it can exploit the dynamic sparsity of the activations. When both are equipped with EIE, the speedup ratio $\eta$ of the model accelerated by RRM compared to the original model can be calculated as:
\vspace{-0.2cm}
\begin{equation}\label{speed-ratio}
\vspace{-0.2cm}
	\eta = \frac{\sum_{i}\rho_{c_i}W_{c_i}H_{c_i}C_{c_i}^{in}C_{c_i}^{out}w_Fh_F+\sum_{j}\rho_{f_j}C_{f_j}^{in}C_{f_j}^{out}}{\sum_{i}\hat{\rho}_{c_i}W_{c_i}H_{c_i}C_{c_i}^{in}C_{c_i}^{out}w_Fh_F+\sum_{j}\hat{\rho}_{f_j}C_{f_j}^{in}C_{f_j}^{out}},
\end{equation}
where $\hat{\rho}_{c_i}$ and $\hat{\rho}_{f_i}$ are the density of the input tensor in our RRM. 

\subsection{Results on the sparsity}\label{sparse}

\renewcommand \arraystretch{1.2}
\begin{table}[!htb]
  \centering
  \small
  \begin{tabular}{cx{1.2cm}x{1.5cm}x{1.5cm}x{1.5cm}}
    \toprule
   Model & Charades & UCF-101 & MERL \\
    \cmidrule(lr){1-1} \cmidrule(lr){2-2} \cmidrule(lr){3-3} \cmidrule(lr){4-4}
    AlexNet~\cite{NIPS2012_4824} & $35.7\%$ & $35.4\%$ & $34.8\%$   \\
    AlexNet + RRM & \bm{$57.5\%$} & \bm{$60.1\%$} & \bm{$71.8\%$} \\
    \cellcolor{gray!15}Improvement & \cellcolor{gray!15}$21.8\%$ & \cellcolor{gray!15}$24.7\%$ & \cellcolor{gray!15}$37.0\%$ \\
    Speedup ratio &$146\%$ & $154\%$ &$211\%$ \\
    \cmidrule(lr){1-1} \cmidrule(lr){2-2} \cmidrule(lr){3-3} \cmidrule(lr){4-4}
    VGG-16~\cite{simonyan2014very} & $50.4\%$ & $51.3\%$ &$53.2\%$   \\
    VGG-16 + RRM & \bm{$66.4\%$} & \bm{$70.1\%$} &\bm{$75.2\%$}   \\
    \cellcolor{gray!15}Improvement & \cellcolor{gray!15} $16.0\%$ & \cellcolor{gray!15} $18.8\%$ & \cellcolor{gray!15}$22.0\%$ \\
    Speedup ratio &$124\%$ &$128\%$ &$136\%$ \\
    \cmidrule(lr){1-1} \cmidrule(lr){2-2} \cmidrule(lr){3-3} \cmidrule(lr){4-4}
    ResNet-18~\cite{he2016deep} & $40.5\%$ &$40.4\%$  &$40.0\%$  \\
    ResNet-18 + RRM & \bm{$58.0\%$} &\bm{$58.4\%$}  &\bm{$73.6\%$}  \\
    \cellcolor{gray!15}Improvement & \cellcolor{gray!15}$17.5\%$ & \cellcolor{gray!15}$18.0\%$ & \cellcolor{gray!15}$33.6\%$ \\
    Speedup ratio &$126\%$&$130\%$ &$191\%$\\
    \bottomrule
  \end{tabular}
  \vspace{-0.3cm}
  \caption{Overall sparsity improvement and the speedup ratio of each model evaluated on three benchmark datasets. Model plus RRM means that we apply our recurrent residual module to the original baseline model. RRM clearly brings significant improvement over the baselines.}
  \vspace{-0.5cm}
  \label{improve-sparse}
\end{table}

To show that our RRM framework is able to generally improve the overall sparsity, we evaluate our method on three different real-time video benchmark datasets: Charades~\cite{sigurdsson2016hollywood}, UCF-101~\cite{soomro2012ucf101}, MERL~\cite{singh2016multi}, and choose three classical deep networks: AlexNet~\cite{NIPS2012_4824}, VGG-16~\cite{simonyan2014very}, ResNet-18~\cite{he2016deep} to be our base networks. In order to formulate the real-time analysis on videos, we sample the video frames at 24 FPS, which is the original frame rate in Charades, and then perform inference that extracts the deep features of these video frames. We measure the overall sparsity improvement of each network when performing inference with our RRM on these three datasets, during which the threshold $\epsilon$ in RRM (as is illustrated in Sec.~\ref{sparse_enhance}) is set to be $10^{-2}$. And the results are recorded in Table~\ref{improve-sparse}. It can be seen that our RRM framework can generally improve the overall sparsity of the input feature maps in DNNs and deliver a speedup as calculated by Eq.~\ref{speed-ratio}. 
This sparsity improvement comparison between datasets indicates that the similarity property of video frames is efficiently exploited by our RRM framework.

\begin{figure}[!htb]
	\begin{center}
	\vspace{-0.2cm}
		\includegraphics[width=0.9\linewidth]{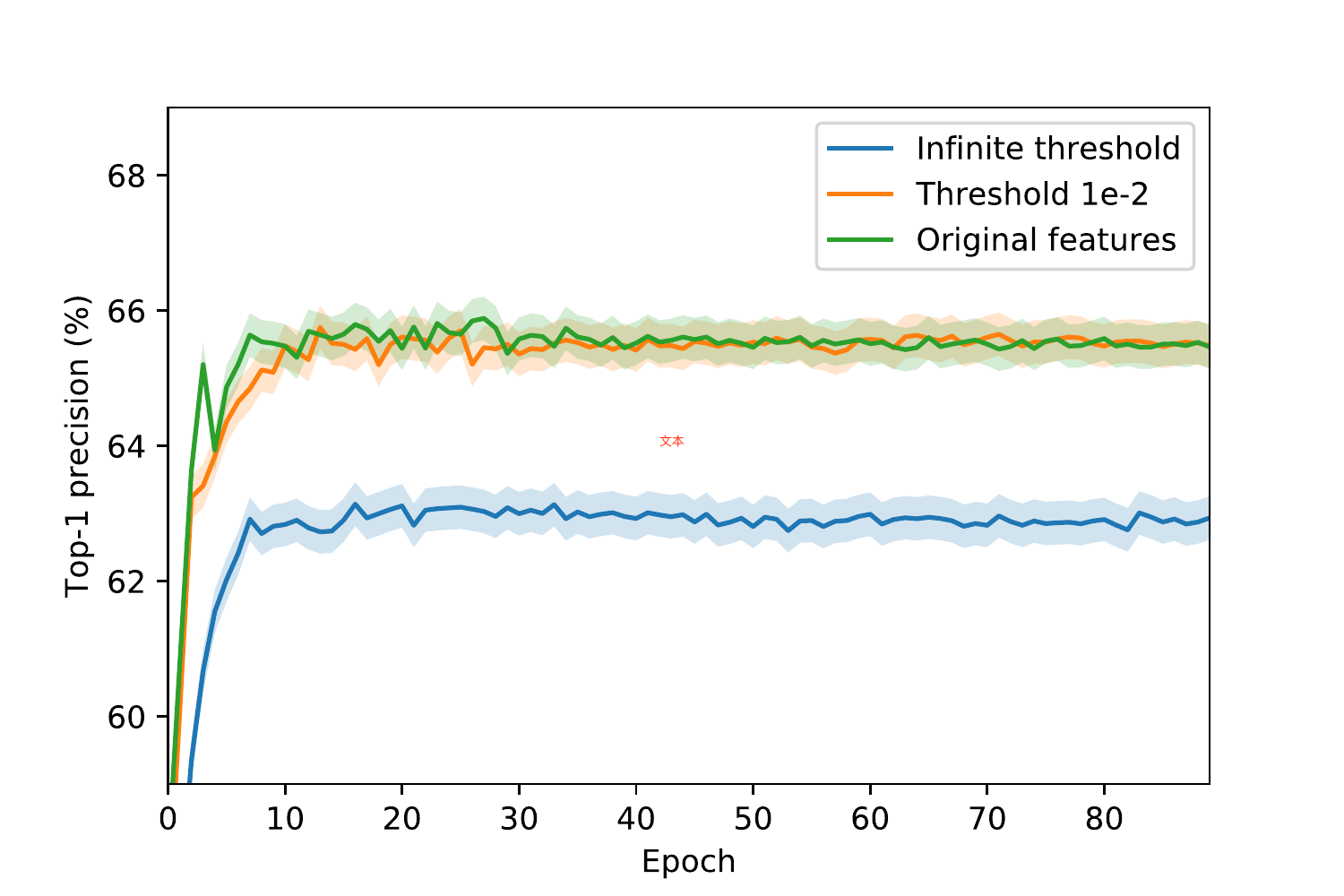}
		\vspace{-0.4cm}
		\caption{We train these features for the action recognition task on UCF-101 (as is discussed in Sec.~\ref{trade-off}). The blue curve represents the performance of the feature extracted with an infinite threshold. We add this ablation curve to show that precisely-computed features are necessary to obtain a good performance.}
		\vspace{-0.7cm}
  		\label{fig:training}
	\end{center}
\end{figure}

Here we also want to clarify the threshold setting. In fact, it makes no difference to treat such small-value elements as zero elements. The $L_2$ distance between the feature extracted under this setting and the original feature is generally around $10^{-6}$. This is a trivial deviation for that, in contrast, translating the cropped image by one pixel can result in an $L_2$ error around $10^{-2}$. As shown in Fig.~\ref{fig:training}, features extracted under this threshold setting have no difference with the original features.

\subsection{Trade-off between accuracy and speedup}\label{trade-off}
In Sec.~\ref{sparse_enhance}, we introduced a sparsity enhancement scheme, which truncates some small values into zero. It can further accelerate the model, but bring some deviation between the calculated feature maps and the original feature maps. Thus, there naturally exists a trade-off between speed and accuracy by adjusting the threshold $\epsilon$. 

\begin{figure}[!htb]
	\begin{center}
		\includegraphics[width=0.9\linewidth]{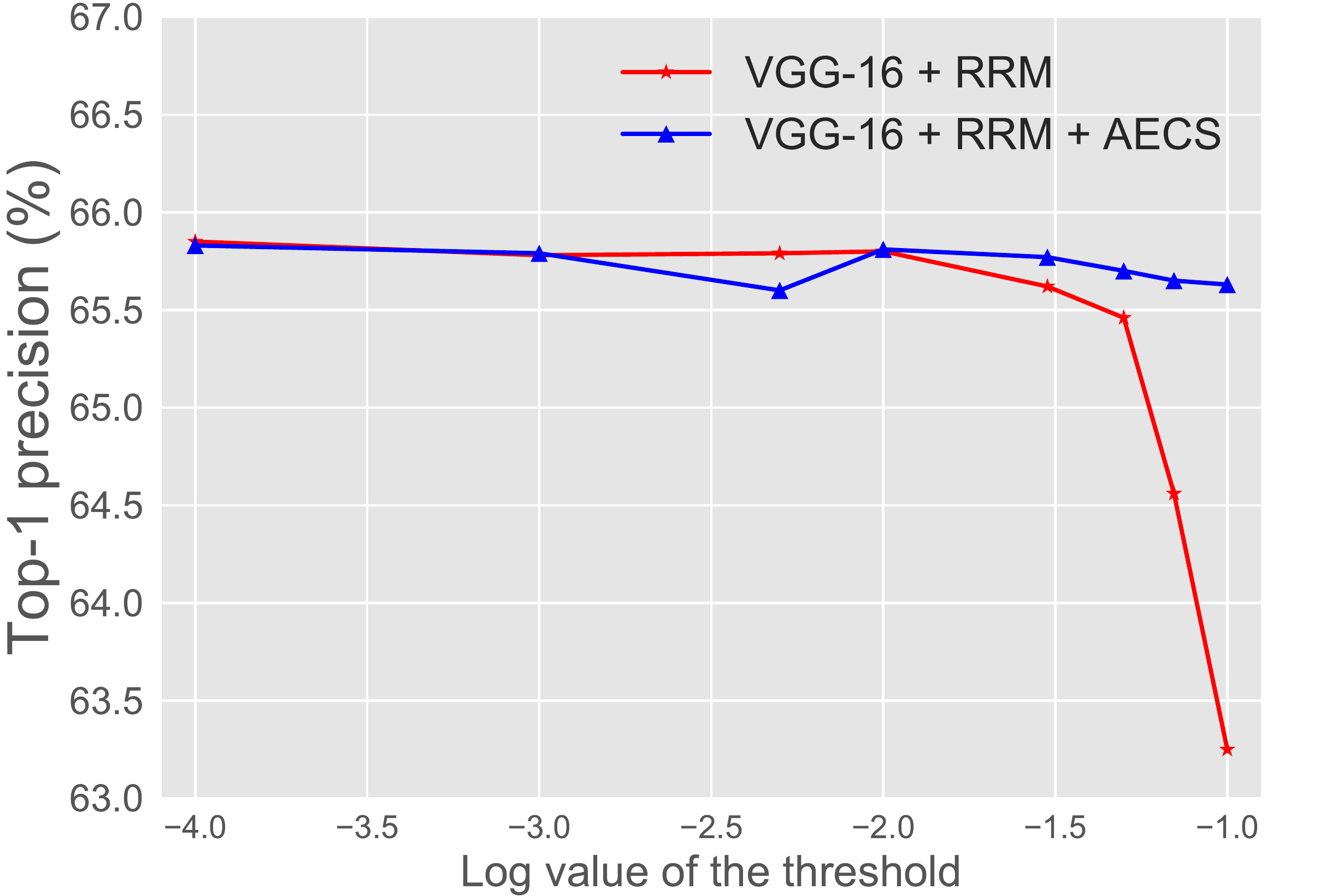}
		\vspace{-0.3cm}
		\caption{We measure the accuracy of features extracted under different threshold settings and validate the effectiveness of the accumulated error control scheme.}
		\vspace{-0.7cm}
  		\label{fig:trade-off}
	\end{center}
\end{figure}
\renewcommand \arraystretch{1.2}
\begin{table}[!htb]
  \centering
  \small
  \begin{tabular}{x{2cm}x{0.8cm}x{1.2cm}x{1.2cm}x{0.8cm}}
    \toprule
   Threshold $\epsilon$ & $10^{-2}$ & $3\times10^{-2}$ & $5\times10^{-2}$ & $10^{-1}$ \\
    \cmidrule(lr){1-1} \cmidrule(lr){2-2} \cmidrule(lr){3-3} \cmidrule(lr){4-4} \cmidrule(lr){5-5}
    Speedup ratio &$1.3\times$&$1.4\times$&$1.6\times$&$1.8\times$ \\
    \bottomrule
  \end{tabular}
  \vspace{-0.2cm}
  \caption{Different speedup ratios of VGG-16 at different level of thresholds.}
  \label{speedup-ratio}
  \vspace{-0.3cm}
\end{table}

We explore this trade-off by performing the action recognition task on UCF-101 dataset~\cite{soomro2012ucf101}. For each video, we first extract the VGG-16 feature vectors of its frames. Then, we perform the average pool on these feature vectors to obtain a video-level feature vector in 4096 dimensions to represent this video. With these video-level features, we train a two-layer MLP to recognize the actions in these videos and evaluate the top-1 precision. As is shown in Fig.~\ref{fig:trade-off}, by gradually amplifying the threshold $\epsilon$ when extracting the feature, the speed up ratio increases while the accuracy drops due to the exploded accumulated error.

\begin{figure}[!htb]
	\begin{center}
	\vspace{-0.3cm}
		\includegraphics[width=0.9\linewidth]{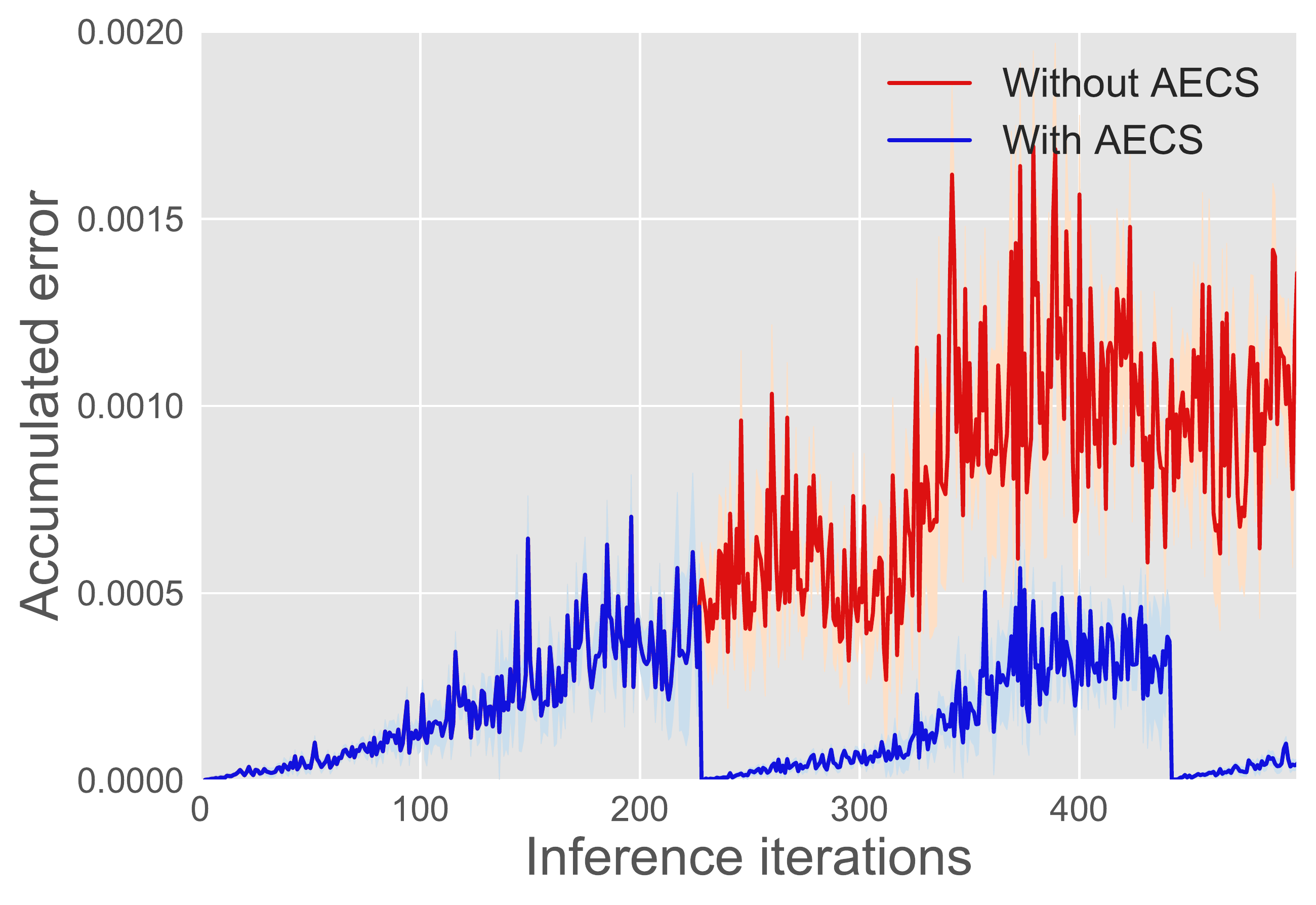}
		\vspace{-0.3cm}
		\caption{The dynamic accumulated error with the threshold $\epsilon$ set as $3\times 10^{-2}$.}
  		\label{fig:erro}
	\vspace{-0.7cm}
	\end{center}
\end{figure}

We then validate the effectiveness of accumulated error control scheme (AECS), which is introduced in Sec.~\ref{sparse_enhance}. With the protection of AECS, the precision is maintained as the $\epsilon$ grows up. Dynamic accumulated error during inference is shown in Fig.~\ref{fig:erro}. We can see that, with a moderate $\epsilon$, the inference speed will not be affected since the expensive original inference is rare. 

\vspace{-0.1cm}
\subsection{Speed up deeply compressed models}\label{compress}

\renewcommand \arraystretch{1.2}\label{compress-speed}
\begin{table}[!htb]
  \centering
  \small
  \vspace{-0cm}
  \begin{tabular}{x{3.4cm}x{1.8cm}x{1.8cm}}
    \toprule
   Model & Charades & MERL \\
    \cmidrule(lr){1-1} \cmidrule(lr){2-2} \cmidrule(lr){3-3}
    Deep Compression & $41.0\%$ & $37.3\%$    \\
    Deep Compression + RRM & $59.4\%$ & $70.2\%$ \\
    \cellcolor{gray!15}Improvement & \cellcolor{gray!15}$18.4\%$ & \cellcolor{gray!15}$32.9\%$\\
    Speedup ratio & \bm{$145\%$} & \bm{$210\%$} \\
    \cmidrule(lr){1-1} \cmidrule(lr){2-2} \cmidrule(lr){3-3}
    XNOR-Net & $0.1\%$ & $0.1\%$\\
    XNOR-Net + RRM & $83.2\%$ & $89.2\%$\\
    \cellcolor{gray!15}Improvement & \cellcolor{gray!15}$83.1\%$ & \cellcolor{gray!15}$89.1\%$\\
    Speedup ratio & \bm{$598\%$} & \bm{$927\%$} \\
    \bottomrule
  \end{tabular}
  \vspace{-0.2cm}
  \caption{Comparison of overall sparsity of XNOR-Nets and deep compression models with RRM.}
  \vspace{-0.4cm}
  \label{compressedmodels}
\end{table}

We examine the performance of RRM on some already-accelerated models and show that these models can be further accelerated by our RRM framework on video inference.

\textbf{Deep compression model.} Han \etal~\cite{han2015deep} proposed the deep compression model, which effectively reduces the model size and the energy consumption. There is a three-stage pipeline that prunes redundant connections between layers, quantizes parameters and compresses model with Huffman encoding. Deep compression model can be largely accelerated in efficient inference engine~\cite{Han:2016:EEI:3007787.3001163}. Efficient inference engine is a general methodology that compresses and accelerates DNNs. We show we can further accelerate the model when processing video frames.

\textbf{XNOR-Net.} Deep CNN models can be sped up by binarizing the input and the weight of the network. Rastegari \etal~\cite{rastegari2016xnor} devised the XNOR-Nets which approximated the original model with binarized input and parameters and achieved a 58$\times$ faster convolution operation. Value of elements in both the input and the weight of the XNOR-Net is transformed to $+1$ or $-1$ by taking their signs. Consequently, convolution operation can be implemented with only additions. The sparsity of feature maps in XNOR-Net is very poor due to the binarization. With RRM applied, the overall sparsity is significantly improved. Besides, after skipping zero-value input elements, the elements remained to be calculated are all $+2$ or $-2$, where the advantages of binary convolution operation can still be maintained by scaling a factor 0.5.

Experiment results can be referred in Table~\ref{compressedmodels}. It demonstrates that our RRM is able to achieve an impressive speedup ratio on these compressed models.

\vspace{-0.15cm}
\subsection{Video pose estimation and object detection}\label{pose-detection}
\begin{figure*}
\begin{center}
\includegraphics[width=\linewidth]{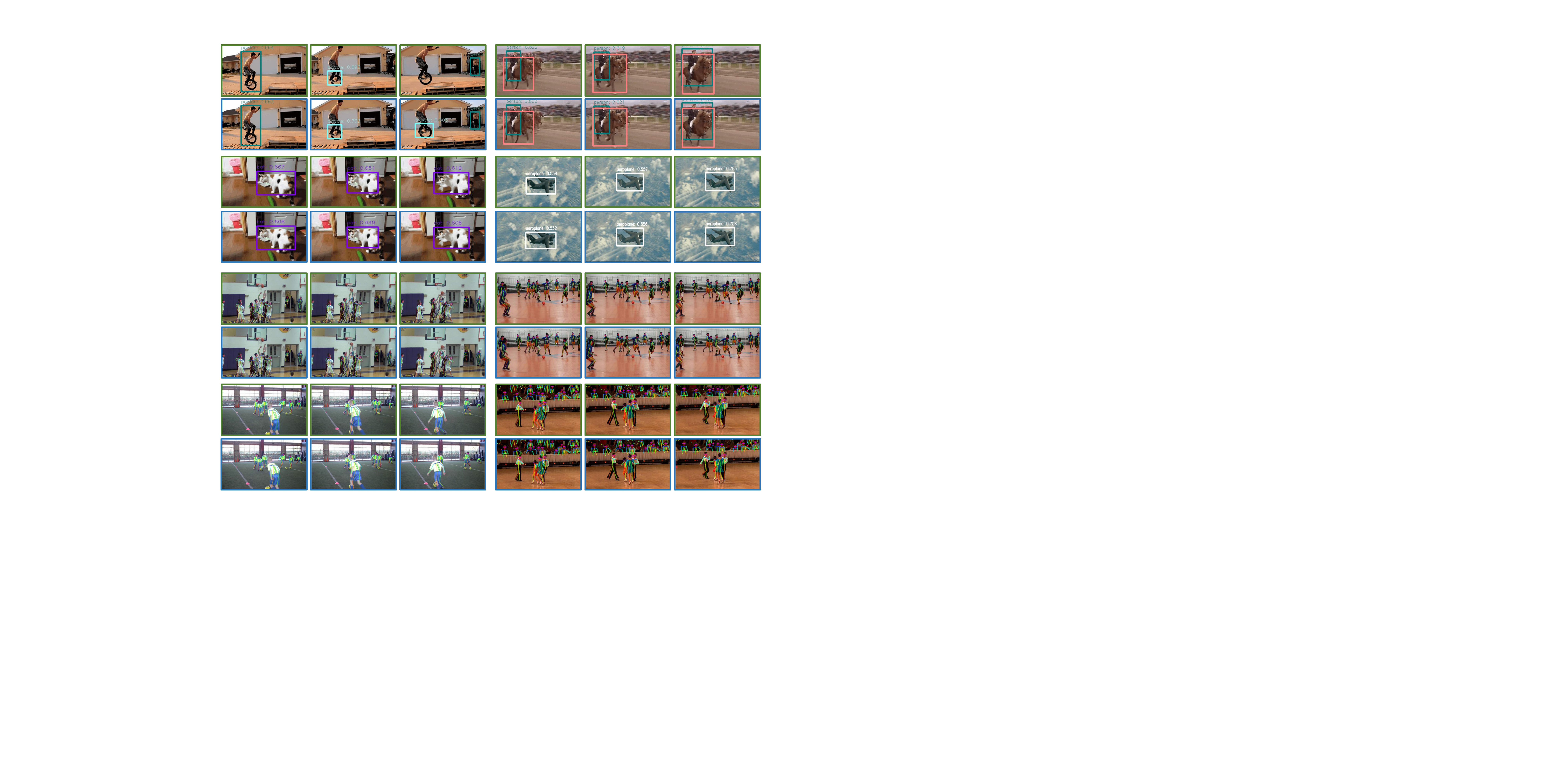}
\end{center}
\vspace{-0.5cm}
\caption{Qualitative results of the object detection and pose estimation in videos. frames with green border are the original results evaluated in a frame-by-frame manner and frames with blue border are the results of our RRM framework. The performance is not affected, and interestingly, we find that our RRM framework sometimes gets a more reliable result than the original model. It could be that our RRM framework can utilize the temporal context information across consecutive frames.}
\vspace{-0.4cm}
\label{samples}
\end{figure*}

\vspace{-0.15cm}
In this section, we apply our RRM framework to several mainstream visual systems to improve the efficiency of their backbone CNN models. We choose two video recognition tasks, video pose estimation and video object detection, to verify the effectiveness of our RRM framework. We set the threshold $\epsilon$ as $10^{-2}$ in the experiments. It is a precise setting which has been validated by preceding experiments in Sec.~\ref{sparse} so that the output features are almost the same as the original model and the recognition performance will not be affected. Some qualitative results are shown in Fig.~\ref{samples}.

\renewcommand \arraystretch{1.2}
\begin{table}[!htb]
  \centering
  \small
  \vspace{-0.3cm}
  \begin{tabular}{x{2.5cm}x{2.5cm}x{2.5cm}}
    \toprule
   Model & MPII Video Pose & BBC Pose \\
    \cmidrule(lr){1-1} \cmidrule(lr){2-2} \cmidrule(lr){3-3}
    rt-Pose\cite{cao2016realtime} & $78.5\%$ & $79.6\%$  \\
    rt-Pose + RRM &  $91.0\%$ & $93.3\%$ \\
    \cellcolor{gray!15}Improvement & \cellcolor{gray!15}$12.5\%$ &  \cellcolor{gray!15}$13.7\%$ \\
    Speedup ratio& \bm{$213.7\%$} & \bm{$291.2\%$} \\
    \bottomrule
  \end{tabular}
  \vspace{-0.25cm}
  \caption{Comparison of the overall sparsity of pose estimator rt-Pose with RRM.}
  \vspace{-0.2cm}
  \label{pose}
\end{table}

\textbf{Video pose estimation.} Real-time video pose estimation is a rising topic in computer vision. To meet the requirement of inference speed, our RRM can be applied for acceleration. Currently, the fastest multi-person pose estimator is the rt-Pose model proposed by Cao \emph{et al.}~\cite{cao2016realtime}, which can reach a speed of 8.8 FPS with one NVIDIA GeForce GTX-1080 GPU. In this part, we apply our RRM framework to further accelerate the rt-Pose model. We evaluate the models on two video pose datasets, BBC Pose\cite{Charles13a} and MPII-Video-Pose~\cite{insafutdinov17arttrack}. The BBC Pose dataset consists of 20 TV broadcast videos (each 0.5h-1.5h in length) while the MPII Video Pose dataset is composed of 28 sequences which contains some challenging frames in the MPII dataset~\cite{andriluka14cvpr}. The experiment results are shown in Table~\ref{pose}, we can see that by applying our RRM, pose estimation in videos are significantly accelerated.

\renewcommand \arraystretch{1.2}
\begin{table}[!htb]
  \centering
  \small
  \begin{tabular}{cx{1.2cm}x{1.8cm}x{1.2cm}x{1.2cm}}
    \toprule
   Model & Charades & UCF-101 & MERL \\
    \cmidrule(lr){1-1} \cmidrule(lr){2-2} \cmidrule(lr){3-3} \cmidrule(lr){4-4}
    YOLOv2\cite{redmon2016yolo9000} & $2.45\%$ & $2.55\%$ & $2.46\%$   \\
    YOLOv2 + RRM & $50.01\%$ & $52.26\%$ & $48.21\%$ \\
    \cellcolor{gray!15}Improvement & \cellcolor{gray!15}$47.56\%$ & \cellcolor{gray!15}$49.71\%$ & \cellcolor{gray!15}$45.75\%$ \\
    Speedup ratio& \bm{$204.9\%$} & \bm{$200.8\%$} & \bm{$197.3\%$} \\
    \bottomrule
  \end{tabular}
  \vspace{-0.3cm}
  \caption{Comparison of overall sparsity of object detection model YOLOv2 with RRM.}
  \vspace{-0.4cm}
  \label{detection}
\end{table}

\textbf{Video object detection.} Majority of the work on object detection is focused on image rather than videos. Redmon \etal~\cite{redmon2016you, redmon2016yolo9000} created YOLO network, which achieved very efficient end-to-end training and testing for object detection. We apply our RRM framework to accelerates the YOLO network to realize a faster real-time detection in videos. We evaluate the models on video object detection on Charades, UCF-101, and MERL. YOLOv2 uses the Leaky-ReLU as the activation function, thus it prevents the sparsity of the original model. By applying our RRM, there brings a huge improvement. As shown in Table~\ref{detection}, the sparsity of original model ranges between $2\%$ and $3\%$. With our RRM, the sparsity increases to $48\%$-$52\%$. In total, our RRM brings a speedup ratio around $200\%$.
\begin{table}[!htb]
\centering
\small
\vspace{-0.2cm}
\begin{tabular}{x{2.1cm}x{0.7cm}x{1.8cm}x{0.68cm}}
	\toprule
   \textsc{Objects} &  mAP  & \textsc{Keypoints} & mAP   \\
	\cmidrule(lr){1-1} \cmidrule(lr){2-2} \cmidrule(lr){3-3} \cmidrule(lr){4-4}  
	YOLOv2 & $61.2$ &  rt-Pose & 46.2  \\
	\cmidrule(lr){1-1} \cmidrule(lr){2-2} \cmidrule(lr){3-3} \cmidrule(lr){4-4} 
	YOLOv2+RRM & \cellcolor{gray!15}$61.1$ &  rt-Pose+RRM & \cellcolor{gray!15} 46.2  \\
	\bottomrule
\end{tabular}	
	\vspace{-0.2cm}
	\caption{Detection and pose estimation performance results.}
    \vspace{-0.3cm}
	\label{perfomance}
\end{table}

\textbf{Recognition accuracy.} To prove that our method is able to maintain performance while greatly accelerate the model inference, we conduct the detection experiments on the Youtube-BB dataset using YOLOv2 and the pose estimation experiments on MPII video pose dataset using rt-Pose. We keep all the training conditions as the same. And the accuracy results are shown in Table~\ref{perfomance}.

\subsection{Discussion}
\begin{figure}[!htb]
	\begin{center}
    \vspace{-0.2cm}
	\includegraphics[width=0.9\linewidth]{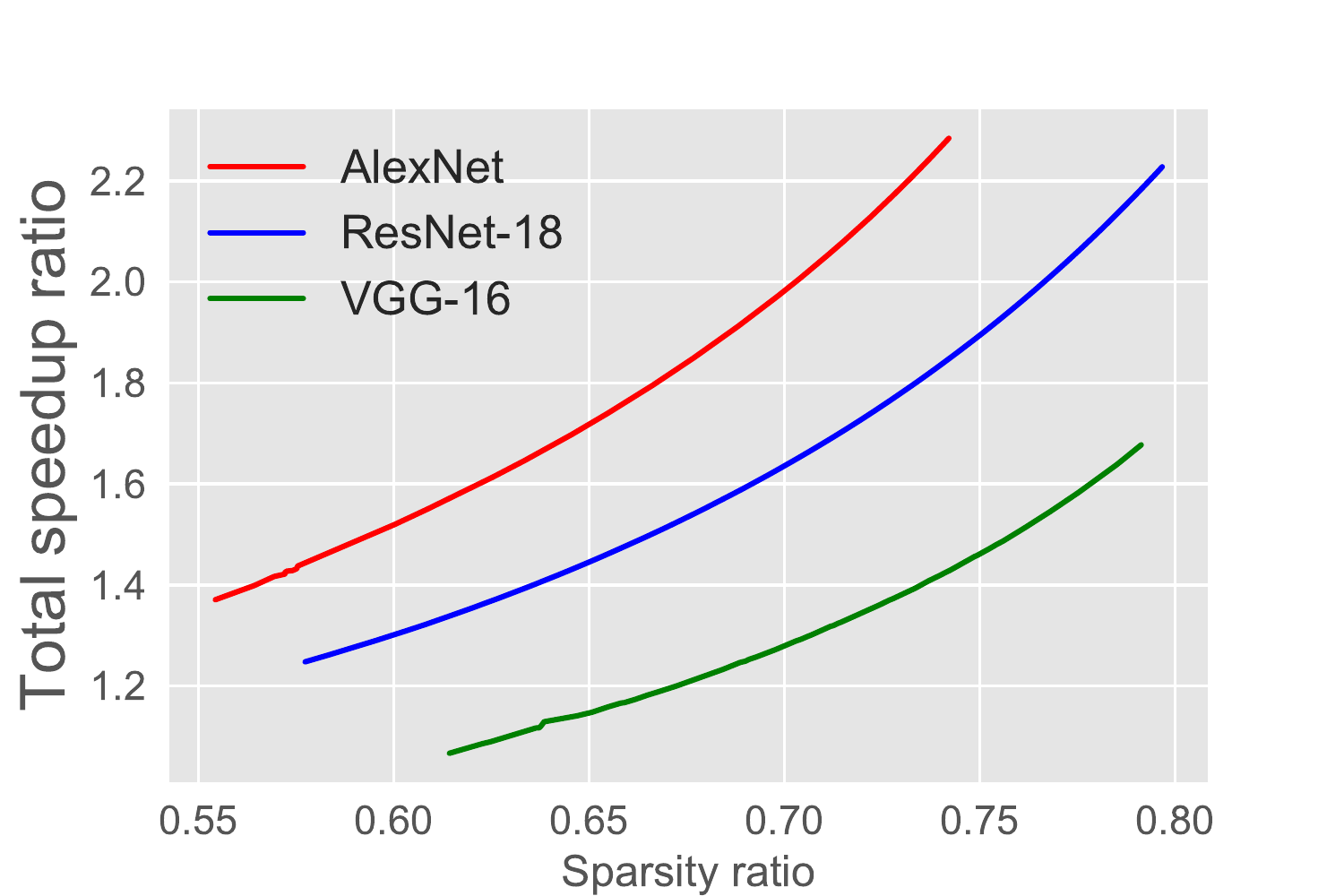}
    \vspace{-0.4cm}
	\caption{Trade-off comparison between speedup and sparsity ratio. We do experiments on the UCF-101 dataset.}
	\vspace{-0.5cm}
    \label{trade-off-total}	
	\end{center}
\end{figure}

\textbf{Theoretical vs. Actual speedup.} Hardware designing to evaluate actual speedup is beyond the scope of the current work, while according to Table III in \cite{Han:2016:EEI:3007787.3001163} actual speedup can be well estimated by the sparsity of weight and activation on EIE engine. It can be seen from Table III in \cite{Han:2016:EEI:3007787.3001163} that the relationship between density of the layer (Weight\%$\times$Act\%) and the speedup of layer inference (FLOP\%) is near-linear. Thus, it can be inferred that, with well-designed hardwares, there won't be a significant performance gap between these theoretical numbers and those in real application.

\textbf{Batch Normalization.} Several studies have shown that the linear layer calculation only occupied part of total inference time, some other non-linear layers are also time-consuming, especially the BN layer. Thus, here we compare the trade-off between total speedup (with all overhead considered) and sparsity ratio among AlexNet (no BN), VGG-16 (no BN) and ResNet-18 (with BN) in Fig.~\ref{trade-off-total}.

\section{Conclusion}
We proposed the Recurrent Residual Module for fast inference in videos. We have shown that the overall sparsity of different CNN models can be generally improved by our RRM framework. Meanwhile, applying our RRM framework to some already-accelerated models, such as XNOR-Net and Deep Compression Model, they can achieve further speedup. Experiments showed that the proposed RRM framework speeds up the visual recognition systems YOLOv2 and rt-Pose for real-time video understanding, delivering impressive speedup without a loss in recognition accuracy. 


{\small
\bibliographystyle{ieee}
\bibliography{egbib}
}

\end{document}